\documentclass[11pt,a4paper]{article}
\usepackage[hyperref]{acl2019}
\usepackage{times}
\usepackage{latexsym}

\usepackage{url}

\aclfinalcopy % Uncomment this line for the final submission
\def\aclpaperid{2369} %  Enter the acl Paper ID here

\usepackage{graphicx}
\usepackage{adjustbox} %package dropping figures
\usepackage{url}
\usepackage[T1]{fontenc}

\usepackage{xspace}
\usepackage{graphicx}
\usepackage{adjustbox}
\usepackage{subcaption}
\usepackage{amsfonts}
\usepackage{color}
\usepackage{multirow}
\usepackage{booktabs}
\usepackage{array}

\newcommand{\BLEU}{{{\textsc{Bleu}}}\xspace}
\newcommand{\NIST}{{{\textsc{Nist}}}\xspace}
\newcommand{\METEOR}{{{\textsc{Meteor}}}\xspace}

\newcommand{\sts}{{{\textsc{Seq2Seq}}}\xspace}
\newcommand{\MemNet}{{{\textsc{MemNet}}}\xspace}

\newcommand{\OurSAN}{{{\textsc{CMR}}}\xspace}
\newcommand{\OurSANw}{{{\textsc{CMR+w}}}\xspace}
\newcommand{\OurSANnf}{{{\textsc{CMR-f}}} \xspace}

\usepackage{enumitem}
\usepackage{amsmath,amsfonts,amssymb,bbm,epsfig,bm}

\usepackage{todonotes}[] % insert [disable] to disable all %Format add comments
\DeclareGraphicsExtensions{.pdf,.png,.jpg}

\author{Lianhui Qin$^\bold{\dagger}$, Michel Galley$^\bold{\ddagger}$, Chris Brockett$^\bold{\ddagger}$, Xiaodong Liu$^\bold{\ddagger}$, \\
{\bf Xiang Gao}$^\bold{\ddagger}$, {\bf Bill Dolan}$^\bold{\ddagger}$, {\bf Yejin Choi}$^\bold{\dagger}$ and {\bf Jianfeng Gao}$^\bold{\ddagger}$ \\
  $^\bold{\dagger}$
  University of Washington, Seattle, WA, USA \\
  $^\bold{\ddagger}$    
  Microsoft Research, Redmond, WA, USA \\
  {\tt\small \{lianhuiq,yejin\}@cs.washington.edu} \\
  {\tt\small \{mgalley,Chris.Brockett,xiaodl,xiag,billdol,jfgao\}@microsoft.com}
}
\date{}

\title{
Conversing by Reading: \\
Contentful Neural Conversation with On-demand Machine Reading}
  
\begin{document}
\maketitle

\begin{abstract}
Although neural conversation models are effective in learning \emph{how} to produce fluent responses, their primary challenge lies in knowing \emph{what} to say to make the conversation \emph{contentful} and non-vacuous. 
We present a new end-to-end approach to contentful neural conversation that jointly models response generation and on-demand machine reading. 
The key idea is to provide the conversation model with relevant long-form text \textit{on the fly} as a source of external knowledge. The model performs QA-style reading comprehension on this text in response to each conversational turn, thereby allowing for more focused integration of external knowledge than has been possible in prior approaches. 
To support further research on knowledge-grounded conversation, we introduce a new large-scale conversation dataset grounded in external web pages (2.8M turns, 7.4M sentences of grounding). Both human evaluation and automated metrics show that our approach results in more contentful responses compared to a variety of previous methods, improving both the informativeness and diversity of generated output.

\end{abstract}

\section{Introduction}
\label{sec:intro}\begin{figure}[!t]
\centering
{\includegraphics[scale=0.39]{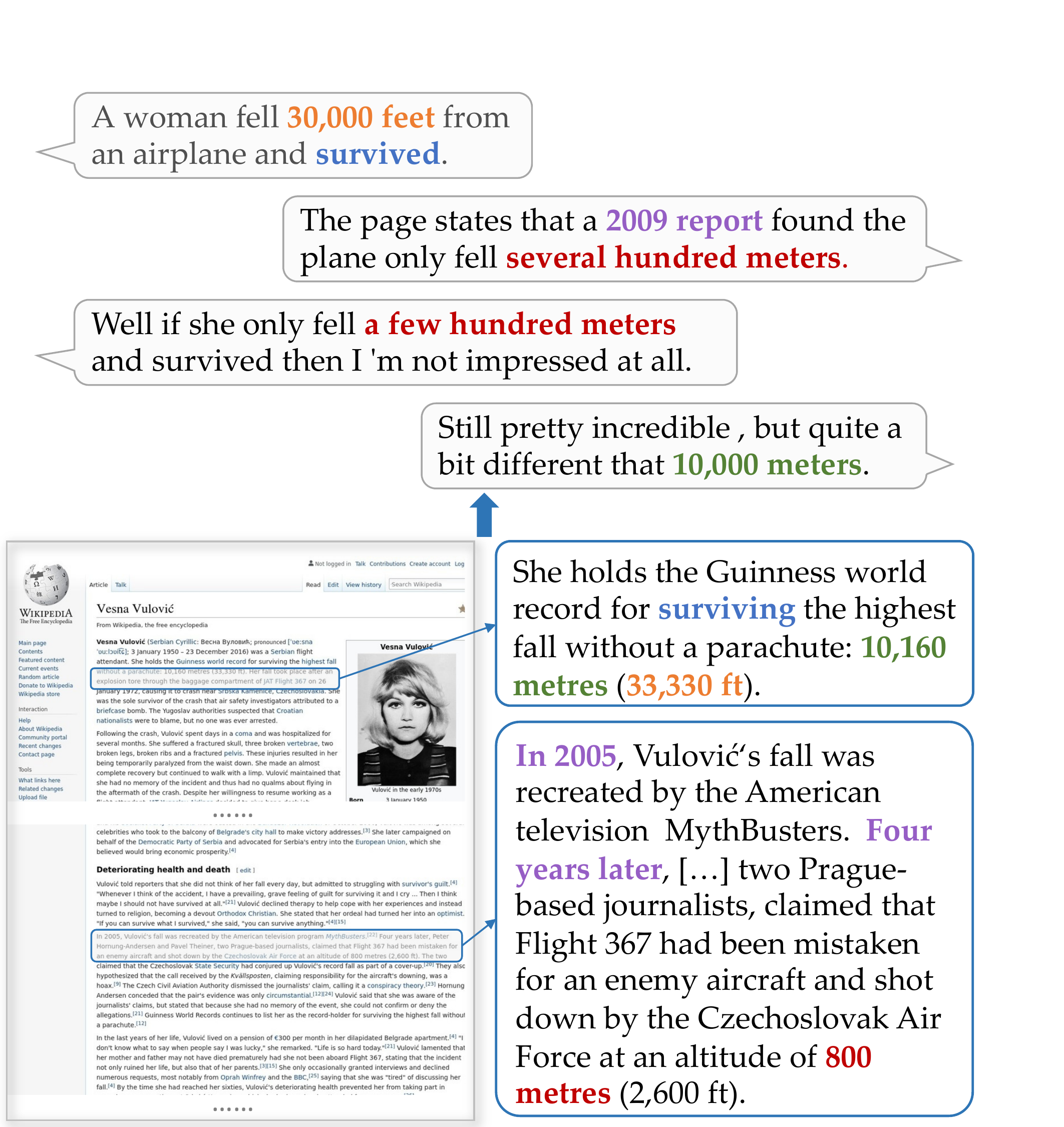}}
\caption{Users discussing a topic defined by a Wikipedia article. In this real-world example from our Reddit dataset, information needed to ground responses is distributed throughout the source document.}
\label{fig:motivation} 
\end{figure}

While end-to-end neural conversation models \cite[etc.]{shang2015neural,sordoni2015neural,vinyals2015neural,serban2015hierarchical,li2016diversity,gao2019neural} are effective in learning \emph{how} to be fluent, their responses are often vacuous and uninformative. A primary challenge thus lies in modeling \emph{what} to say to make the conversation contentful. 
Several recent approaches have attempted to address this difficulty by conditioning the language decoder on external information sources, such as knowledge bases~\cite{agarwal:18,liu2018knowledge}, review posts~\cite{marjan18,moghe2018towards}, and even images~\cite{das2017visual,mostafazadeh2017image}.
However, empirical results suggest that conditioning the decoder on rich and complex contexts, while helpful, does not on its own provide sufficient inductive bias for these systems to learn how to achieve deep and accurate integration between external knowledge and response generation.

We posit that this ongoing challenge demands a more  
effective mechanism to support on-demand knowledge integration. We draw inspiration from how humans converse about a topic, where people often search and acquire external information as needed to continue a meaningful and informative conversation. Figure~\ref{fig:motivation} illustrates an example human discussion, where information scattered in separate paragraphs must be consolidated to compose grounded and appropriate responses. 
Thus, the challenge is to {connect the dots} across different pieces of information in much the same way that \emph{machine reading comprehension (MRC)} systems tie together multiple text segments to provide a unified and factual answer \cite[etc.]{Seo2016BidirectionalAF}. 

We introduce a new framework of end-to-end conversation models that jointly learn response generation together with on-demand machine reading. 
We formulate the reading comprehension task as document-grounded response generation: given a long document that supplements the conversation topic, along with the conversation history, we aim to produce a response that is both conversationally appropriate and informed by the content of the document. The key idea is to project conventional QA-based reading comprehension onto conversation response generation by equating the conversation prompt with the question, the conversation response with the answer, and external knowledge with the context.
The MRC framing allows for integration of long external documents that present notably richer and more complex information than relatively small collections of short, independent review posts such as those that have been used in prior work~\citep{marjan18,moghe2018towards}. %

We also introduce a large dataset to facilitate research on knowledge-grounded conversation (2.8M turns, 7.4M sentences of grounding) that is at least one order of magnitude larger than existing datasets~\citep{conversationalwikipedia,moghe2018towards}.
This dataset consists of real-world conversations extracted from Reddit, linked to web documents discussed in the conversations.
Empirical results on our new dataset demonstrate that our full model improves over previous grounded response generation systems and various ungrounded baselines, suggesting that deep knowledge integration is an important 
research direction.\footnote{Code for reproducing our models and data is made publicly available at \url{https://github.com/qkaren/converse_reading_cmr}.} 
\section{Task}
\label{sec:task}We propose to use factoid- and entity-rich web documents, e.g., news stories and Wikipedia pages, as external knowledge sources for an open-ended conversational system to ground in.

Formally, we are given a conversation history of turns $X = (\bm{x}_1,\dots,\bm{x}_M)$ and a web document \mbox{$D=(\bm{s}_1,\dots,\bm{s}_N)$} as the knowledge source, where $\bm{s}_i$ is the $i$th sentence in the document. 
With the pair $(X, D)$, the system needs to generate a natural language response $\bm{y}$ that is both conversationally appropriate and reflective of the contents of the web document.
\section{Approach}
\label{sec:model}\begin{figure*}[t]
\centering
{\includegraphics[scale=0.45]{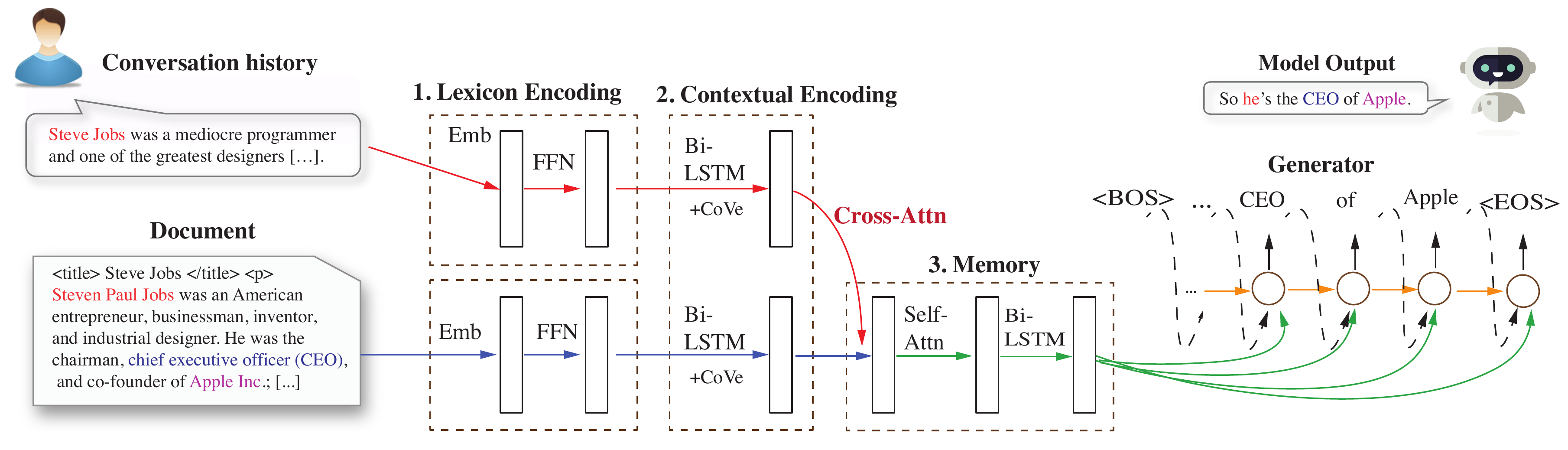}}
\caption{{\bf Model Architecture for Response Generation with on-demand Machine Reading:}
The first blocks of the MRC-based encoder serve as a lexicon encoding that maps words to their embeddings and transforms with position-wise FFN, independently for the conversation history and the document. 
The next block is for contextual encoding, where BiLSTMs are applied to the lexicon embeddings to model the context for both conversation history and document. 
The last block builds the final encoder memory, by sequentially applying cross-attention in order to integrate the two information sources, conversation history and document, self-attention for salient information retrieval, and a BiLSTM for final information rearrangement. 
The response generator then attends to the memory and generates a free-form response.}
\label{fig:model} 
\end{figure*}

Our approach integrates conversation generation with on-demand MRC.
Specifically, we use an MRC model to effectively encode the conversation history by treating it as a question in a typical QA task (e.g., SQuAD~\citep{rajpurkar2016squad}), and encode the web document as the context. We then replace the output component of the MRC model (which is usually an answer classification module) with an attentional sequence generator that generates a free-form response. We refer to our approach as CMR (\underline{C}onversation with on-demand \underline{M}achine \underline{R}eading). In general, any off-the-shelf MRC model could be applied here for knowledge comprehension. We use Stochastic Answer Networks (SAN)\footnote{\url{https://github.com/kevinduh/san_mrc}}
\citep{liu2017stochastic}, 
a performant machine reading model that until very recently held state-of-the-art performance on the SQuAD benchmark.
We also employ a simple but effective data weighting scheme to further encourage response grounding.

\subsection{Document and Conversation Reading}

We adapt the SAN model to encode both the input document and conversation history and forward the digested information to a response generator. 
Figure~\ref{fig:model} depicts the overall MRC architecture. 
Different blocks capture different concepts of representations in both the input conversation history and web document. 
The leftmost blocks represent the lexicon encoding that extracts information from $X$ and $D$ at the token level. Each token is first transformed into its corresponding word embedding vector, and then fed into a position-wise feed-forward network (FFN) \cite{vaswani2017attention} to obtain the final token-level representation. Separate FFNs are used for the conversation history and the web document.

The next block is for contextual encoding. The aforementioned token vectors are concatenated with pre-trained 600-dimensional CoVe vectors~\cite{mccann2017learned},
and then fed to a BiLSTM that is shared for both conversation history and web document. The step-wise outputs of the BiLSTM carry the information of the tokens
as well as their left and right context.

The last block builds the memory that summarizes the salient information from both $X$ and $D$. The block first applies \emph{cross}-attention to integrate information from the conversation history $X$ into the document representation. Each contextual vector of the document $D$ is used to compute attention (similarity) distribution over the contextual vectors of $X$, which is concatenated with the weighted average vector of $X$ by the resulting distribution. Second, a \emph{self}-attention layer is applied to further ingest and capture the most salient information. The output memory, $M\in \mathbb{R}^{d \times n}$, is obtained by applying another BiLSTM layer for final information rearrangement. Note that $d$ is the hidden size of the memory and $n$ is the length of the document.

\subsection{Response Generation}
Having read and processed both the conversation history and the extra knowledge in the document, 
the model then produces a free-form response ${\bf y} = (y_1,\dots,y_T)$ 
instead of generating a span or performing answer classification as in MRC tasks.

We use an attentional recurrent neural network decoder~\citep{luong2015effective} to generate response tokens while attending to the memory. At the beginning, the initial hidden state $\bm{h}_0$ is the weighted sum of the representation of the history $X$. 
For each decoding step $t$ with a hidden state $\bm{h}_t$, we generate a token $y_t$ based on the distribution:
\begin{equation}
p(y_t) = \text{softmax}((W_1 \bm{h}_t + \bm{b})/\tau ),
    \label{eq:gen}
\end{equation}
where $\tau > 0$ is the softmax temperature.
The hidden state $\bm{h}_t$ is defined as follows:
\begin{equation}
\bm{h}_t = W_2[\bm{z}_t +\!\!+ f_{attention}(\bm{z}_t, M)].
    \label{eq:hidden_t}
\end{equation}
Here, $[\cdot +\!\!+  \cdot]$ indicates a concatenation of two vectors; $f_{attention}$ is a dot-product attention \cite{vaswani2017attention}; and $\bm{z}_t$ is a state generated by $\text{GRU}(\bm{e}_{t-1}, \bm{h}_{t-1})$
with $\bm{e}_{t-1}$ being the embedding of the word $y_{t-1}$ generated at the previous ($t-1$) step. In practice, we use top-$k$ sample decoding to draw $y_t$ from the above distribution $p(y_t)$. 
Section~\ref{sec:exp} provides more details
about the experimental configuration.

\subsection{Data Weighting Scheme}
We further propose a simple data weighting scheme to encourage the generation of grounded responses. The idea is to bias the model training to fit better to those training instances where the 
ground-truth 
response is more closely relevant to the document. More specifically, given a training instance $(X, D, \bm{y})$, we measure the closeness score $c\in\mathbb{R}$ between the document $D$ and the 
gold response $\bm{y}$ (e.g., with the NIST~\citep{Doddington:2002} or \BLEU~\citep{papineni2002bleu} metrics). In each training data batch, we normalize the closeness scores of all the instances to have a sum of $1$, and weight each of the instances with its corresponding normalized score when evaluating the training loss. This training regime promotes instances with grounded responses and thus encourages the model to better encode and utilize the information in the document. 
\section{Dataset}
\label{sec:dataset}
\begin{table}
\small
    \centering
    \begin{tabular}{rlll}
    \cmidrule[\heavyrulewidth]{1-4}
      & Train & Valid & Test \\ 
    \cmidrule{1-4}
    \# dialogues    & 28.4k   & 1.2k   & 3.1k \\
    \# utterances   & 2.36M    & 0.12M   & 0.34M \\
    \# documents     & 28.4k   & 1.2k   & 3.1k \\ %
    \# document sentences & 15.18M &  0.58M &  1.68M \\
    \cmidrule{1-4}
    \multicolumn{3}{l}{\it Average length (\# words):}  \\
    utterances & 18.74 & 18.84 & 18.48 \\
    document sentences & 13.72 & 14.17  & 14.15  \\
    \cmidrule[\heavyrulewidth]{1-4}
    \end{tabular}
    \caption{Our grounded conversational dataset.} 
    \label{tab:stats}
\end{table}

To create a grounded conversational dataset,
we extract conversation threads from Reddit, %
a popular and large-scale online platform for news and discussion. In 2015 alone, 
Reddit hosted more than 73M conversations.\footnote{\url{https://redditblog.com/2015/12/31/reddit-in-2015/}}
On Reddit, user submissions are categorized by topics or ``subreddits'', and a submission typically consists of a submission title associated with a URL pointing to a news or background article, which initiates a discussion about the contents of the article. This article provides framing for the conversation, and this can naturally be seen as a form of grounding. Another factor that makes Reddit conversations particularly well-suited for our conversation-as-MRC setting is that a significant proportion of these URLs contain named anchors (i.e., `\#' in the URL) that point to the relevant passages in the document.
This is conceptually quite similar to MRC data~\citep{rajpurkar2016squad} where typically only short passages within a larger document are relevant in answering the question.

We reduce spamming and offensive language by manually curating a list of 178 relatively ``safe'' subreddits and 226 web domains from which the web pages are extracted. 
To convert the web page of each conversation into a text document, we extracted the text of the page using an html-to-text converter,\footnote{\url{https://www.crummy.com/software/BeautifulSoup}} 
while retaining important tags such as \textsf{\textless title\textgreater}, \textsf{\textless h1\textgreater} to \textsf{\textless h6\textgreater}, and \textsf{\textless p\textgreater}. 
This means the entire text of the original web page is preserved, but these main tags 
retain some high-level structure of the article.
For web URLs with named anchors, we preserve that information by indicating the anchor text in the document with tags \textsf{\textless anchor\textgreater} and \textsf{\textless /anchor\textgreater}. As the whole documents in the dataset tend to be lengthy, anchors offer important hints to the model about which parts of the documents should likely be focused on in order to produce a good response. We considered it sensible to keep them as they are also available to the human reader.

After filtering short or redacted turns, or which quote earlier turns, we obtained 2.8M conversation instances respectively divided into train, validation, and test (Table~\ref{tab:stats}). We used different date ranges for these different sets: years 2011-2016 for train, Jan-Mar 2017 for validation, and the rest of 2017 for test. For the test set, we select conversational turns for which 6 or more responses were available, in order to create a multi-reference test set. Given other filtering criteria such as turn length, this yields a 6-reference test set of size 2208. For each instance, we set aside one of the 6~human responses to assess human performance on this task, and the remaining 5 responses serve as ground truths for evaluating different systems.\footnote{While this is already large for a grounded dataset, we could have easily created a much bigger one given how abundant Reddit data is. We focused instead on filtering out spamming and offensive language, in order to strike a good balance between data quality and size.}
Table~\ref{tab:stats} provides statistics for our dataset, and 
Figure~\ref{fig:motivation} presents an example from our dataset that also demonstrates the need to combine conversation history and background information from the document to produce an informative response.

To enable reproducibility of our experiments, we crawled web pages using Common Crawl (\url{http://commoncrawl.org}), a service that crawls web pages and makes its historical crawls available to the public. We also release the code (URL redacted for anonymity) to recreate our dataset from both a popular Reddit dump\footnote{\url{http://files.pushshift.io/reddit/}} and Common Crawl, and 
the latter service ensures that anyone reproducing our data extraction experiments would retrieve exactly the same web pages.
We made a preliminary version of this dataset available for a shared task \cite{galley:19} at Dialog System Technology Challenges (DSTC) \cite{dstc7}. Back-and-forth with participants helped us iteratively refine the dataset. The code to recreate this dataset is included.\footnote{We do not report on shared task systems here, as these systems do not represent our work and some of these systems have no corresponding publications. Along with the data described here, we provided a standard \sts baseline to the shared task, which we improved for the purpose of this paper (improved \BLEU, \NIST and \METEOR). Our new \sts baseline is described in Section~\ref{sec:exp}.}
\section{Experiments}
\label{sec:exp}\subsection{Systems}

\begin{table*}[t]
\small
	\centering
	\begin{tabular}{@{}r | l l l| l l l| l l l| l@{}}
	\cmidrule[\heavyrulewidth]{1-11}
	& \multicolumn{3}{c|}{{\bf Appropriateness}} & \multicolumn{3}{c|}{{\bf Grounding}} & \multicolumn{3}{c|}{{\bf Diversity}} &
	\\\cmidrule{1-11}
	  & \NIST & \BLEU & \METEOR  & Precision & Recall & F1 & Entropy-4 & Distinct-1 & Distinct-2 & Len \\ 
		\cmidrule{1-11}
Human &2.650&3.13\%&8.31\%&2.89\%&0.45\%&0.78\% &10.445&0.167&0.670&18.757\\
    \cmidrule{1-11}
\sts{} & 2.223 & 1.09\% & 7.34\% & 1.20\% & 0.05\% & 0.10\% & 9.745 & 0.023 & 0.174 & 15.942 \\
\MemNet{} & 2.185 & 1.10\% & 7.31\% & 1.25\% &0.06\% &0.12\% & 9.821 & 0.035 & 0.226 & 15.524 \\
	\cmidrule{1-11}
\OurSANnf{}  & {\bf2.260} & 1.20\% & 7.37\% &1.68\%	& 0.08\%	& 0.15\% & 9.778 & 0.035 & 0.219 & 15.471 \\
\OurSAN{} & 2.213 & {\bf1.43}\% & 7.33\% & 2.44\% & 0.13\%	& 0.25\% & 9.818 & 0.046 & 0.258 & 15.048 \\
\OurSANw{} & 2.238 & 1.38\% & {\bf7.46}\% & {\bf3.39}\% & {\bf0.20}\% & {\bf0.38}\% & {\bf9.887} & {\bf0.052} & {\bf0.283} & 15.249 \\
	\cmidrule[\heavyrulewidth]{1-11}
	\end{tabular}
	\caption{{\bf Automatic Evaluation} results (higher is better for all metrics). Our best models (\OurSANw and \OurSAN) considerably increase the quantitative measures of Grounding, and also slightly improve Diversity. Automatic measures of Quality (e.g., \BLEU-4) give mixed results, but this is reflective of the fact that we did not aim to improve response relevance with respect to the context, but instead its level of grounding.
	The human evaluation results in Table~\ref{tab:humanevalresults} 
	indeed suggest that our best system (\OurSANw) is better.}
    \label{tab:results}
\end{table*}

We evaluate our systems and several competitive baselines:\\
{\bf \sts}~\citep{sutskever2014sequence}
We use a standard LSTM \sts model that only exploit the conversation history for response generation, without any grounding. 
This is a competitive baseline
initialized using pretrained embeddings.\\
{\bf \MemNet{}:} We use a Memory Network %
designed for grounded response generation~\citep{marjan18}.
An end-to-end memory network~\citep{sukhbaatar2015end} encodes conversation history and sentences in the web documents. Responses are generated with a sequence decoder.\\
{\bf \OurSANnf{}}: To directly measure the effect of incorporating web documents, we compare to a baseline which omits the document reading component of the full model (Figure~\ref{fig:model}). As with the \sts approach, the resulting model generates responses solely based on conversation history.\\
{\bf \OurSAN{}}:
To measure the effect of our data weighting scheme, we compare to a system that has identical architecture to the full model, but is trained without associating weights to training instances.\\
{\bf \OurSANw{}}: As described in section~\ref{sec:model}, the full model reads and comprehends both the conversation history and document using an MRC component, and sequentially generates the response.  The model is trained with the data weighting scheme to encourage grounded responses.\\
{\bf Human}:
To get a better sense of the systems' performance relative to an upper bound,
we also evaluate human-written responses using different metrics. As described in Section~\ref{sec:dataset}, for each test instance, we set aside one of the 6 human references for evaluation, so the `human' is evaluated against the other 5 references for automatic evaluation. To make these results comparable, all the systems are also automatically evaluated against the same 5 references.

\section{Experiment Details}

For all the systems, we set word embedding dimension to 300 and used the pretrained GloVe\footnote{\url{https://nlp.stanford.edu/projects/glove/}} for initialization. 
We set hidden dimensions to 512 and dropout rate to 0.4. GRU cells are used for \sts and \MemNet (we also tested LSTM cells and obtained similar results). We used the Adam optimizer for model training, with an initial learning rate of 0.0005. Batch size was set to 32. During training, all responses were truncated to have a maximum length of 30, and maximum query length and document length were set to 30, 500, respectively. we used regular teacher-forcing decoding during training. For inference, we found that top-$k$ random sample decoding~\cite{fan2018hierarchical} provides the best results for all the systems. That is, at each decoding step, a token was drawn from the $k$ most likely candidates according to the distribution over the vocabulary. Similar to recent work~\citep{fan2018hierarchical,edunov2018understanding}, we set $k=20$ (other common $k$ values like 10 gave similar results). We selected key hyperparameter configurations on the validation set. %

\subsection{Evaluation Setup}

Table~\ref{tab:results} shows automatic metrics for quantitative evaluation over three qualities of generated texts.
We measure the overall {\bf relevance} of the generated responses given the conversational history by using standard Machine Translation (MT) metrics, comparing generated outputs to ground-truth responses. These metrics include
\BLEU-4~\citep{papineni2002bleu}, 
\METEOR~\citep{Lavie2007}. 
and \NIST~\citep{Doddington:2002}.
The latter metric is a variant of \BLEU that weights $n$-gram matches by their information gain by effectively penalizing uninformative $n$-grams (such as ``I don't know''), 
which makes it a relevant metric for evaluating systems aiming diverse and informative responses. 
MT metrics may not be particularly adequate for our task \cite{liu2016how}, given its focus on the informativeness of responses, and for that reason we also use two other types of metrics to measure the level of grounding and diversity.

As a {\bf diversity} metric,
we count all $n$-grams in the system output for the test set, and measure: (1) Entropy-$n$ as the entropy of the $n$-gram count distribution, a metric proposed in \cite{zhang18towards}; (2) Distinct-$n$ as the ratio between the number of $n$-gram types and the total number of $n$-grams, a metric introduced in \cite{li2016diversity}.

For the {\bf grounding} metrics, 
we first compute `\#match,' the number of non-stopword tokens in the response that are present in the document but not present in the context of the conversation. Excluding words from the conversation history means that, in order to produce a word of the document, the response generation system is very likely to be effectively influenced by that document. We then compute both \emph{precision} as `\#match' divided by the total number of non-stop tokens in the response, and \emph{recall} as `\#match' divided by the total number of non-stop tokens in the document. We also compute the respective \emph{F1} score to combine both.
Looking only at exact unigram matches between the document and response is a major simplifying assumption, but the combination of the three metrics offers a plausible proxy for how greatly the response is grounded in the document. It seems further reasonable to assume that these can serve as a surrogate for less quantifiable forms of grounding such as paraphrase -- e.g., {\it US} $\xrightarrow{}$ {\it American} -- when the statistics are aggregated on a large test dataset.

\subsection{Automatic Evaluation}

Table~\ref{tab:results} shows automatic evaluation results for the different systems. %
In terms of appropriateness, the different variants of our models outperform the \sts and \MemNet baselines, but differences are relatively small and, in case of one of the metrics (\NIST), the best system does not use grounding. %
Our goal, we would note, is not to specifically improve response appropriateness, as many responses that completely ignore the document (e.g., {\it I don't know}) might be perfectly appropriate. Our systems fare much better in terms of Grounding and Diversity: our best system (\OurSANw) achieves an F1 score that is more than three times (0.38\% vs. 0.12\%) higher than the most competitive non-MRC system (\MemNet). 

\begin{table}[t]
\small
\centering
\begin{tabular}{r@{\hskip7pt}r|r|r@{\hskip10pt}l}
\cmidrule[\heavyrulewidth]{1-5}
\multicolumn{5}{c}{{\it Human judges preferred:}}\\[0.1cm]

\multicolumn{2}{c|}{Our best system} & Neutral & \multicolumn{2}{c}{Comparator} \\ 
\cmidrule{1-5}
\OurSANw &  *{\bf44.17}\% &  26.27\% &  29.56\% &  \sts \\
\OurSANw &  *{\bf 40.93}\% &  25.80\% &  33.27\% &  \MemNet \\
\OurSANw &   {\bf37.67}\% &  27.53\% &  34.80\% &  \OurSAN\\
\cmidrule{1-5}
\OurSANw & 30.37\%  &  16.27\% &  *{\bf 53.37}\% &  Human \\
	\cmidrule[\heavyrulewidth]{1-5}
	\end{tabular}
\caption{{\bf Human Evaluation} results, showing preferences (\%) for our model (\OurSANw) vs. baseline and other comparison systems. Distributions are skewed towards \OurSANw. The 5-point Likert scale has been collapsed to a 3-point scale. *Differences in mean preferences are statistically significant (p  $\leq$ 0.0001).}\label{tab:humanevalresults}
\end{table}

\subsection{Human Evaluation}
\label{sec:human_eval}

We sampled 1000 conversations from the test set. Filters were applied to remove conversations containing ethnic slurs or other offensive content that might confound judgments. Outputs from systems to be compared were presented pairwise to judges from a crowdsourcing service. Four judges were asked to compare each pair of outputs on Relevance (the extent to which the content was related to and appropriate to the conversation)  and Informativeness (the extent to which the output was interesting and informative). Judges were asked to agree or disagree with a statement that one of the pair was better than the other on the above two parameters, using a 5-point Likert scale.\footnote{The choices presented to the judges were \textit{Strongly Agree}, \textit{Agree}, \textit{Neutral}, \textit{Disagree}, and \textit{Strongly Disagree}.} Pairs of system outputs were randomly presented to the judges in random order in the context of short snippets of the background text. 
These results are presented in summary form in Table~\ref{tab:humanevalresults}, which shows the overall preferences for the two systems expressed as a percentage of all judgments made. 
Overall inter-rater agreement measured by Fliess' Kappa was 0.32 (``fair"). Nevertheless, the differences between the paired model outputs are statistically significant (computed using 10,000 bootstrap replications).

\subsection{Qualitative Study}

Table~\ref{fig:examples} illustrates how
our best model (\OurSANw) tends to produce more contentful and informative responses compared to the other systems. In the first example, our system refers to a particular {\it episode} mentioned in the article, and also uses terminology that is more consistent with the article (e.g., {\it series}). In the second example, {\it humorous song} seems to positively influence the response, which is helpful as the input doesn't mention singing at all. In the third example, the \OurSANw model clearly grounds its response to the article as it states the fact (Steve Jobs: CEO of Apple) retrieved from the article. The outputs by the other two baseline models are instead not relevant in the context.

\begin{figure}[!t]
\centering
{\includegraphics[scale=0.4]{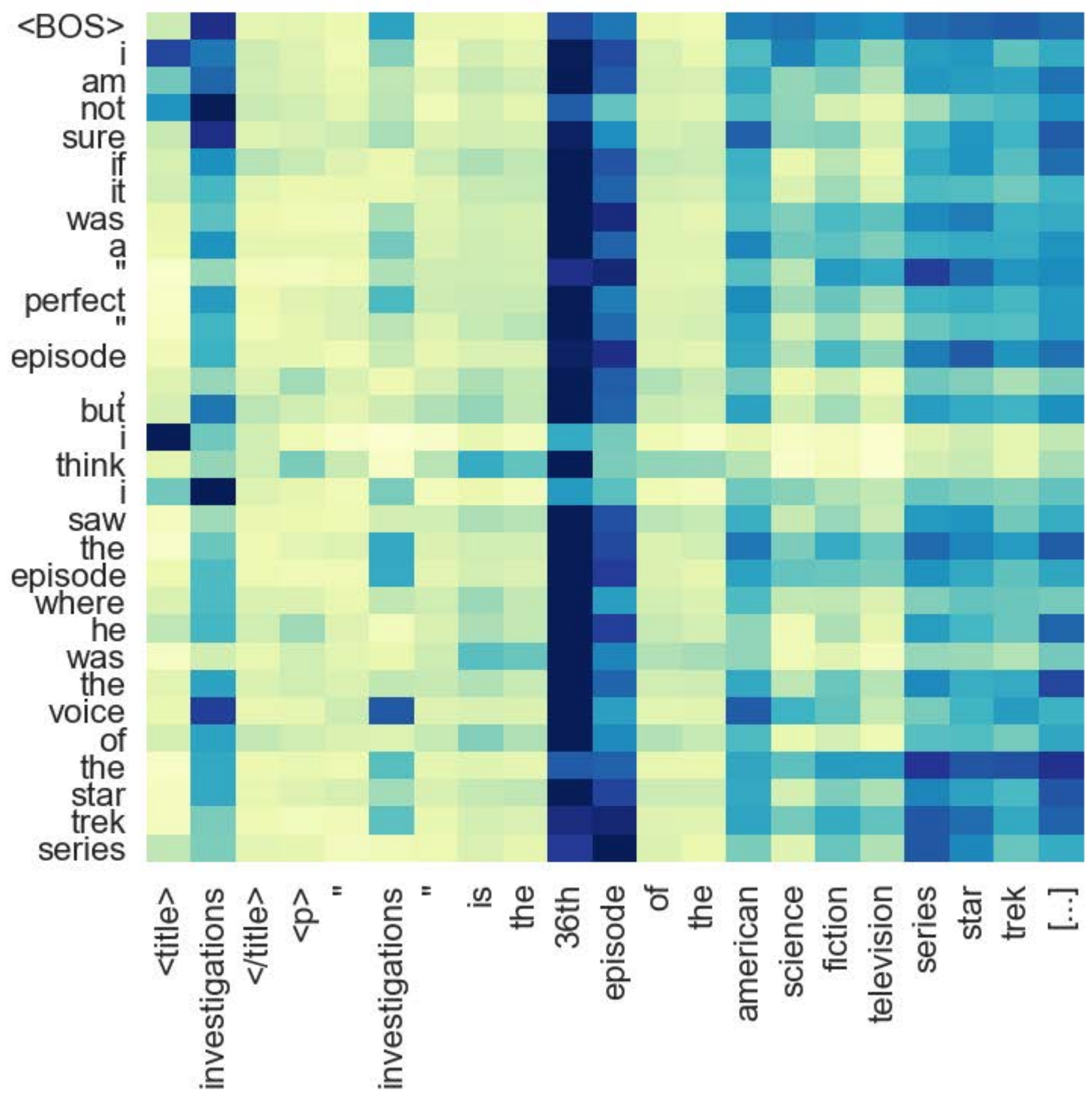}}
\vspace{-10pt}
\caption{Attention weights between words of the documents and words of the response. Dark (blue) cells represent probabilities closer to 1.
}
\label{fig:heatmap} 
\vspace{-10pt}
\end{figure}

Figure~\ref{fig:heatmap} displays the attention map of the generated response and (part of) the document from our full model. The model 
successfully attends to the key words (e.g., {\it 36th, episode}) of the document.
Note that the attention map 
is unlike what is typical in machine translation, where 
target words tend to
attend to different portions of the input text. 
In our task, 
where alignments are much less one-to-one compared to machine translation,
it is common for the generator to retain focus
on the key information in the external document to produce semantically relevant responses.

\begin{table*}[ht]
\centering
\begin{small}
\begin{tabular}{rp{5.4in}}
\toprule
Doc & <title> Investigations </title> <p> "Investigations" is the {\bf 36th episode} of the American science fiction {\bf television series} Star Trek: Voyager which aired on the UPN network. It is the {\bf 36th episode} of the second season. </p> Set in the 24th century, {\bf the series} follows the adventures of the Federation starship Voyager during its journey home to Earth, having been stranded tens of thousands of light-years away. [...]\\
Input & [...] King Abdullah II of Jordon once cameo'd as an unnamed ensign on Star Trek Voyager\\
\MemNet{} &I think it should be noted that he is like a good guy who just killed her and her son. \\
\sts{} &well , i didn't know , but i was just saying that it wasn't a joke. \\ 
\OurSANw & I am not sure if it was a {\bf ``perfect'' episode}, but I think I saw the {\bf episode} where he was the voice of the {\bf Star Trek series}.\\ 
\midrule
Doc & <title> ``Weird Al'' Yankovic </title> <p> Alfred Matthew ``Weird Al'' Yankovic is an American singer, songwriter, parodist, record producer, satirist, actor, voice actor, music video director, film producer, and author. He is {\bf known} for {\bf his humorous songs} that make light of popular culture and often parody specific songs by contemporary musical acts, original songs that are style pastiches of the work of other acts, and polka medleys of several popular songs, featuring his favored instrument, the accordion.</p> {{\it ({\bf Prince} and {\bf West} mentioned several times in the rest of the article.)}} [...]\\
Input & That's fair, I believe Weird Al always got permission from artists before he parodied them which is nice.\\
\MemNet & Well then you don't need to make a sequel to the front page of the movie.\\
\sts & If I remember correctly. I had no idea who had any good chance in history class.\\ 
\OurSANw & No, because {\bf his song was a parody} of the artist {\bf formerly known as "Kanye West"}.\\ 
\midrule
Doc & <title> Steve Jobs </title> <p> Steven Paul Jobs was an American entrepreneur, businessman, inventor, and industrial designer. He was the chairman, {\bf chief executive officer (CEO)}, and co-founder of {\bf Apple Inc}.; chairman and majority shareholder of Pixar; a member of The Walt Disney Company's board of directors following its acquisition of Pixar; and the founder, chairman, and CEO of next. [...] </p>\\
Input & Steve Jobs was a mediocre programmer and one of the greatest designers and marketers in human history. But I prefer Bill Gates if we want to talk about hero worship myself. [...]\\
\MemNet & What if you don't want to see this movie?\\
\sts & No ... the whole point is that this is a good way to make a difference.\\ 
\OurSANw & So he's the {\bf CEO} of {\bf Apple}.\\ 
\bottomrule
\end{tabular}
\end{small}
\caption{Sample output comparing our best system (\OurSANw) against Memory Networks %
and a \sts baseline. 
The source documents were manually shortened to fit in the table, without significantly affecting meaning.}
\label{fig:examples}
\end{table*}
\section{Related Work}
\paragraph{Dialogue:} Traditional dialogue systems (see \cite{jurafsky2009speech} for an historical perspective) are typically grounded, enabling these systems to be reflective of the user's environment. 
The lack of grounding has been a stumbling block for the earliest end-to-end dialogue systems, as various researchers have noted that their outputs tend to be bland \cite{li2016diversity, gao2019spacefusion}, inconsistent \cite{zhang18perso,li2016persona, zhang2019consistent}, and lacking in factual content \cite{marjan18,agarwal:18}. 
Recently there has been growing interest in exploring different forms of grounding, including images, knowledge bases, and plain texts \cite{das2017visual,mostafazadeh2017image,agarwal:18,yang2019hybrid}. A recent survey is included in \citet{gao2019neural}.

Prior work, e.g, \citep{marjan18,zhang18perso,huang2019challenges}, uses grounding in the form of independent snippets of text: Foursquare tips and background information about a given speaker. Our notion of grounding is different, as our inputs are  much richer, encompassing the full text of a web page and its underlying structure.
Our setting also differs significantly from relatively recent  work~\citep{conversationalwikipedia,moghe2018towards} exploiting crowdsourced conversations with detailed grounding labels: we use Reddit because of its very large scale and better characterization of real-world conversations.
We also require the system to learn grounding directly
from conversation and document pairs, instead
of relying on 
additional grounding labels. 
\citet{moghe2018towards} explored directly using a span-prediction QA model for conversation. Our framework differs in that we combine MRC models with a sequence generator to produce free-form responses.

\paragraph{Machine Reading Comprehension:} MRC models such as SQuAD-like models,
aim to extract answer spans (starting and ending indices) from a given document for a given question \cite{Seo2016BidirectionalAF,liu2017stochastic,yu2018qanet}. These models differ in how they fuse information between questions and documents. 
We chose SAN \cite{liu2017stochastic} because of its representative architecture and competitive performance on existing MRC tasks. We note that other off-the-shelf MRC models, such as BERT~\citep{BERT}, can also be plugged in. We leave the study of different MRC architectures for future work.
Questions are treated as entirely independent in these ``single-turn'' MRC models, so
recent work (e.g., CoQA \cite{reddy2018coqa} and QuAC \cite{choi2018quac}) focuses on multi-turn MRC, modeling sequences of questions and answers in a conversation. While multi-turn MRC aims to answer complex questions, that body of work is restricted to factual questions, whereas our work---like much of the prior work in end-to-end dialogue---models free-form dialogue, which also encompasses chitchat and non-factual responses.
\section{Conclusions}
We have demonstrated that the machine reading comprehension approach offers a promising step to generating, \textit{on the fly}, contentful conversation exchanges that are grounded in extended text corpora. The functional combination of MRC and neural attention mechanisms offers visible gains over several strong baselines. We have also formally introduced a large dataset that opens up interesting challenges for future research. 

The CMR (Conversation with on-demand machine reading) model presented here will help connect the many dots across multiple data sources. One obvious future line of investigation will be to explore the effect of other off-the-shelf machine reading models such as BERT \cite{BERT} within the CMR framework.

\section*{Acknowledgements}
We are grateful to the anonymous reviewers, as well as to
Vighnesh Shiv,
Yizhe Zhang, 
Chris Quirk, 
Shrimai Prabhumoye, and
Ziyu Yao
for helpful comments and suggestions on this work. This research was supported in part by NSF (IIS- 1524371), DARPA CwC through ARO (W911NF- 15-1-0543), and Samsung AI Research.

\bibliographystyle{acl_natbib}
\bibliography{main}

\clearpage
\appendix

\end{document}